\documentclass[11pt]{article}
\usepackage[final]{acl}
\usepackage{times}
\usepackage{latexsym}
\usepackage[T1]{fontenc}
\usepackage[utf8]{inputenc}
\usepackage{microtype}
\usepackage{inconsolata}
\usepackage{graphicx}
\usepackage{xspace}
\usepackage{booktabs}
\usepackage{tabularx}
\usepackage{array}
\usepackage{pifont}
\usepackage{amsmath}
\usepackage[inline]{enumitem}
\usepackage{mdframed}
\usepackage{xcolor}
\usepackage{listings}
\usepackage[most]{tcolorbox}

\newcommand{\ourmodel}{\texttt{MARCH}\xspace}

\newcommand\boxedname{Prompt\xspace}

\definecolor{graybg}{HTML}{f8f8f8}
\definecolor{titlebg}{HTML}{e8e8e8}
\definecolor{framecolor}{HTML}{cccccc}
\newtcolorbox[auto counter]{prompt}[1][]{
    breakable,
    enhanced,
    colback=graybg,
    colframe=framecolor,
    colbacktitle=titlebg,
    coltitle=black,
    fonttitle=\scriptsize\bfseries,
    fontupper=\scriptsize\ttfamily,
    title={\boxedname~\thetcbcounter: #1},
    boxrule=0.5pt,
    arc=3pt,
    top=4pt,
    bottom=2pt,
    left=2pt,
    right=2pt,
    before skip=5pt,
    after skip=5pt,
    parbox=true,
}

\title{MARCH: Multi-Agent Radiology Clinical Hierarchy for CT Report Generation}

\author{
 \textbf{Yi Lin\textsuperscript{1}},
 \textbf{Yihao Ding\textsuperscript{2}},
 \textbf{Yonghui Wu\textsuperscript{3}},
 \textbf{Yifan Peng\textsuperscript{1}}
\\
 \textsuperscript{1}Weill Cornell Medicine, New York, USA\\
 \textsuperscript{2}University of Western Australia, Crawley, Australia\\
 \textsuperscript{3}University of Florida, Florida, USA
\\
 \small{
   \textbf{Correspondence:} \href{yip4002@med.cornell.edu}{yip4002@med.cornell.edu}
 }
}

\begin{document}
\maketitle
\begin{abstract}
Automated 3D radiology report generation often suffers from clinical hallucinations and a lack of the iterative verification found in human practice. While recent Vision-Language Models (VLMs) have advanced the field, they typically operate as monolithic "black-box" systems without the collaborative oversight characteristic of clinical workflows. To address these challenges, we propose \ourmodel (\textbf{M}ulti-\textbf{A}gent \textbf{R}adiology \textbf{C}linical \textbf{H}ierarchy), a multi-agent framework that emulates the professional hierarchy of radiology departments and assigns specialized roles to distinct agents. \ourmodel utilizes a \textit{Resident Agent} for initial drafting with multi-scale CT feature extraction, multiple \textit{Fellow Agents} for retrieval-augmented revision, and an \textit{Attending Agent} that orchestrates an iterative, stance-based consensus discourse to resolve diagnostic discrepancies. On the RadGenome-ChestCT dataset, \ourmodel significantly outperforms state-of-the-art baselines in both clinical fidelity and linguistic accuracy. Our work demonstrates that modeling human-like organizational structures enhances the reliability of AI in high-stakes medical domains.
\end{abstract}
\section{Introduction}
\label{sec:intro}

\begin{figure*}[t] 
    \centering    
    \includegraphics[width=\textwidth]{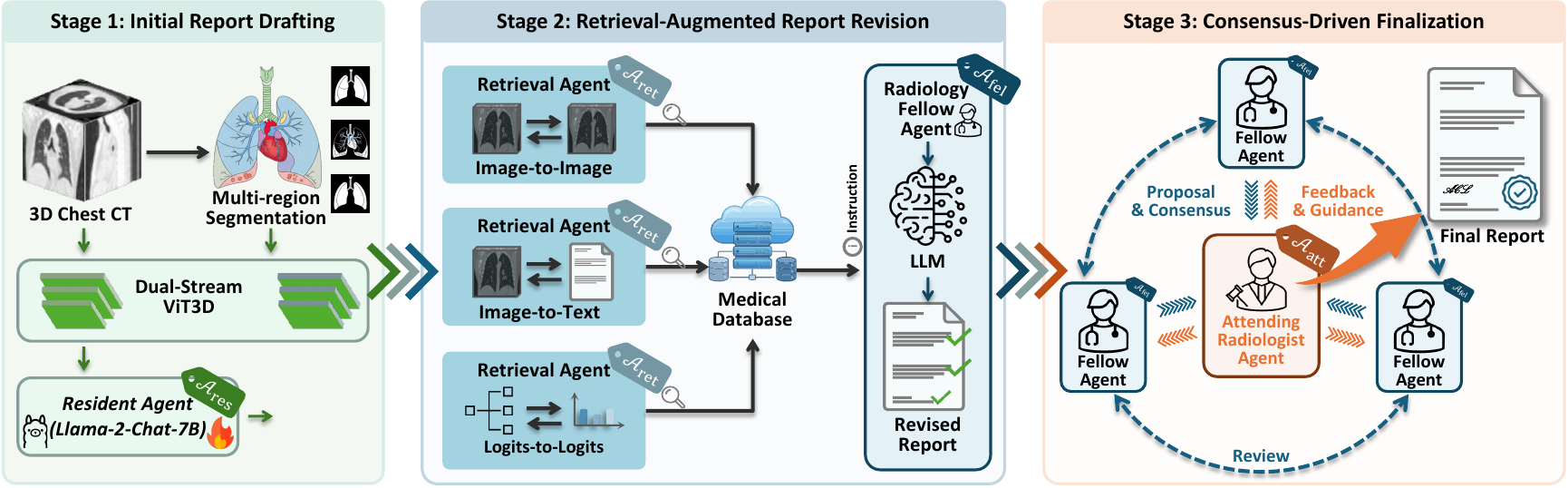}  
    \caption{Overview of the \ourmodel framework. It consists of three main stages: 1) Initial Report Drafting, 2) Retrieval-Augmented Report Revision, and 3) Consensus-Driven Finalization.} 
    \label{fig:overview} 
\end{figure*}

The interpretation of medical imaging, particularly three-dimensional (3D) volumetric data like chest Computed Tomography (CT), remains a cornerstone of modern diagnostic medicine~\cite{moor2023foundation}. 
Despite its importance, generating radiology reports that are accurate, comprehensive, and clinically valid remains cognitively demanding and represents a primary bottleneck in clinical workflows. 
While Large Language Models (LLMs) and Vision-Language Models have shown promise in automating radiology report generation~\cite{ma2025fully}, these approaches often exhibit clinical \textit{hallucinations}, struggle to detect subtle pathological findings in sparse 3D data, and lack the iterative verification and cross-checking~\cite{zhu2025can}.

In radiology, a well-established strategy to reduce such interpretive errors is to provide structured cognitive support via a \textit{devil's advocate} role, commonly implemented through \textit{overread} or \textit{read-out} sessions~\cite{seah2021effect, waite2017interpretive}. Typically, a \textit{Resident} performs the initial interpretation and drafts a report, which is then independently reviewed and, if necessary, reinterpreted by a \textit{Fellow}. When discrepancies persist, an \textit{Attending} radiologist conducts a parallel read to adjudicate and finalize the report. This hierarchical, verification-driven workflow improves diagnostic accuracy of the final report, supports continuous learning through targeted feedback, and enables timely updates to patient management~\cite{hill2017can}. By contrast, most existing automated report generation systems do not model this multi-agent review process, and instead rely on end-to-end, black-box generation~\cite{wang2023metransformer}.

To bridge this gap, we propose \ourmodel (\textbf{M}ulti-\textbf{A}gent \textbf{R}adiology \textbf{C}linical \textbf{H}ierarchy), a consensus-driven multi-agent framework that explicitly models the hierarchical and collaborative structure of radiology \textit{read-out} sessions~\cite{du2025multi-agent, liao2025agentmaster, dang2025multi-agent}. 
Specifically, \ourmodel framework is organized into three layers: 
\begin{enumerate*}[label=(\arabic*)]
\item \textit{Resident Agent} and \textit{Fellow Agent} that generates a detailed initial draft from volumetric scans utilizing a 3D vision encoder coupled with a multi-region segmentation module; 
\item \textit{Retrieval Agents} and \textit{Fellow Agents} that jointly refine the draft by grounding it in evidence-based findings retrieved from an expansive clinical database through multi-modal similarity search; and
\item \textit{Attending Agent} that moderates an iterative consensus-driven discourse. 
\end{enumerate*}

Unlike conventional sequence-to-sequence approaches that merely average outputs, \ourmodel simulates multi-round ``clinical meetings'' in which agents explicitly agree, correct, or refine interpretations until a clinically coherent consensus is reached, thereby naturally aligning with the cognitive structure of decision-making and verification in radiology~\cite{goold2006ethics,bolin2006strategies}.

Our primary contributions are summarized as follows:
\begin{enumerate*}[nosep, label=(\roman*)]
\item We present \ourmodel, a hierarchical multi-agent framework that explicitly models the Resident-Fellow-Attending workflow for automated radiology report generation.
\item We introduce a retrieval-augmented revision stage that utilizes image-, text-, and logit-based indices to improve diagnostic grounding, coherence, and clinical fidelity.
\item We design a consensus-driven finalization protocol in which multiple agents iteratively resolve diagnostic discrepancies through stance-based discourse.
\item Comprehensive empirical analyses and ablation studies demonstrate that \ourmodel significantly outperforms state-of-the-art baselines on both reference-based and clinical validity metrics.
\end{enumerate*}
\section{Related Work}
\label{sec:related_work}

LLMs have demonstrated remarkable capabilities in various medical applications~\cite{singhal2025toward, goh2024large,wang2025accelerating, yamamoto2024enhancing, ma2025fully,liu2024bootstrapping}.
Notably, general-purpose models such as GPT-4~\cite{openai2024gpt4technicalreport} have been shown to outperform medical students on standardized medical exams.
Building on this success, recent studies have explored the use of LLMs for generating medical reports from imaging data~\cite{sloan2024automated}.
These approaches typically involve fine-tuning LLMs with paired image-text datasets~\cite{johnson2019mimic} or employing prompt engineering techniques to employ large pre-trained models~\cite{wang2023r2gengpt}.
These methods have achieved promising results in zero-shot or few-shot settings~\cite{liu2024context}.
However, these methods continue to face challenges such as diverse medical image modalities, complex medical terminology, and clinically unsafe hallucinations~\cite{liu2025argus,jiang2025comt}.

Multi-agent systems extend LLMs by decomposing complex tasks into smaller, manageable sub-tasks~\cite{khan2025comprehensive,plaat2025agentic}.
In the medical domain, such systems have been applied to medical diagnosis~\cite{fan2025ai,kim2024mdagents}, treatment recommendation~\cite{chen2025self}, and medical image analysis~\cite{li2024mmedagent}.
These systems typically integrate task planning, knowledge retrieval, and response generation components to enable structured reasoning and improved reliability~\cite{zhong2025vision,yi2025multimodal}.
Despite these advances, the application of multi-agent systems for 3D radiology report generation remains underexplored.

In contrast to these approaches, \ourmodel addresses the inherent complexity of 3D radiology report generation by distributing heterogeneous tasks across a multi-agent hierarchy equipped with both specialized medical tools (e.g., trainable models) and reasoning capabilities (LLMs). We pivot from monolithic models to a modular, multi-agent framework in which agents adjudicate conflicting findings through dynamic, multi-round negotiation. This design not only enhances interpretability and reliability but also enables seamless integration of domain-specific knowledge, thereby setting a new benchmark in 3D report generation.
\section{Methodology}
\label{sec:method}

The proposed \ourmodel framework operates through three interrelated phases: (1) Initial Report Drafting, (2) Retrieval-Augmented Report Revision, and (3) Consensus-Driven Finalization (Figure~\ref{fig:overview}).

\subsection{Initial Report Drafting}
In this stage, we instantiate a \textit{Resident Agent} $\mathcal{A}_{\text{res}}$ to generate a preliminary radiology report draft $\mathcal{T}$ from chest CT scans $\mathcal{I}$.
The agent is trained on a large-scale corpus of paired volumetric CT scans and reports to learn the cross-modal alignment between visual pathology and textual descriptions.

To mitigate the sparsity of abnormal findings in volumetric data, $\mathcal{A}_{\text{res}}$ embeds a multi-region segmentation module based on the SAT (Segment Anything with Text) model~\cite{zhao2025large-vocabulary}. This module partitions $\mathcal{I}$ into ten anatomical subregions (e.g., bone, breast), allowing the encoder to attend to localized anatomical and pathological entities.

Formally, the report is generated as
$
\mathcal{T} = \mathcal{A}_{\text{res}}(\mathcal{I}; \theta_{\text{res}})
$,
where $\theta_{\text{res}}$ are the learned parameters. Our implementation utilizes a frozen dual-stream ViT3D backbone pre-trained on RadFM~\cite{wu2025towards} for spatial feature extraction and LlaMA-2-Chat-7B~\cite{touvron2023llama} optimized via LoRA~\cite{hu2022lora} for text generation.

\subsection{Retrieval-Augmented Report Revision}
To mitigate omissions and hallucinations, the \textit{Retrieval Agent} $\mathcal{A}_{\text{ret}}$ identifies relevant clinical context from a training database $\mathcal{D}$. We propose three retrieval paradigms:
\begin{enumerate*}[label=(\roman*)]
    \item \textbf{Image-to-Image} \& \textbf{Image-to-Text retrieval}, which uses a 3D vision encoder to retrieve visually similar CT volumes and their corresponding reports from $\mathcal{D}$.
    \item \textbf{Logits-based retrieval}, where a classification head atop $\mathcal{A}_{\text{res}}$ predicts 18 canonical clinical abnormalities (e.g., pleural effusion, atelectasis), and these logits are used to retrieve reports with similar diagnostic profiles.
\end{enumerate*}
In this paper, each retrieval agent retrieves top-3 cases and concatenates them into a structured retrieved evidence $\mathcal{R} = \mathcal{A}_{\text{ret}}(\mathcal{I}, \mathcal{D})$ and then provided to a \textit{Fellow Agent} $\mathcal{A}_{\text{fel}}$, which refines the initial draft by validating findings and resolving inconsistencies, producing an enhanced report $\mathcal{T}'$: 
$
\mathcal{T}' = \mathcal{A}_{\text{fel}}(\mathcal{T}, \mathcal{R})
$.

\subsection{Consensus-Driven Finalization}
The final stage employs a multi-round collaborative protocol orchestrated by an \textit{Attending Agent} $\mathcal{A}_{\text{att}}$ to ensure that the final report reaches a clinical consensus among multiple specialized fellow agents $\{\mathcal{A}_{\text{fel}, i}\}_{i=1}^N$, where $N$ is the number of fellows.

\paragraph{Round 1: Consensus Synthesis.} $\mathcal{A}_{\text{att}}$ first aggregates the enhanced reports from all fellows to generate an initial consensus report $\mathcal{T}^{(0)}$ and identifies potential clinical conflicts:
\begin{equation}
\mathcal{T}^{(0)} = \mathcal{A}_{\text{att}}(\{ \mathcal{T}'_i \}_{i=1}^N).
\end{equation}

\paragraph{Round $t+1$: Iterative Refinement.} In subsequent rounds, each fellow $\mathcal{A}_{\text{fel},i}$ reviews the current consensus $\mathcal{T}^{(t)}$ and provides a stance $S_i^{(t)}$, indicating agreement, proposing corrections, or adding supplementary observations:
\begin{equation}
S_i^{(t)} = \mathcal{A}_{\text{fel}, i}(\mathcal{T}'_i, \mathcal{T}^{(t)}).
\end{equation}
The attending agent $\mathcal{A}_{\text{att}}$ then ensembles these stances to update the report:
\begin{equation}
\mathcal{T}^{(t+1)} = \mathcal{A}_{\text{att}}(\mathcal{T}^{(t)}, \{ S_i^{(t)} \}_{i=1}^N).
\end{equation}

This iteration continues until the attending agent determines that a stable consensus or a predefined maximum number of rounds $T$ has been reached.

\section{Experimental Setting}
\begin{table*}[ht]
\small
\centering
\setlength{\tabcolsep}{4pt}
\begin{tabular}{l*{9}{c}}
\toprule
{Method} & {BLEU-1} & {BLEU-2} & {BLEU-3} & {BLEU-4} & {METEOR} & {ROUGE-L} & {CE-Precision} & {CE-Recall} & {CE-F1}\\ 
\midrule
R2GenPT \shortcite{wang2023r2gengpt} & 0.433 & 0.341 & 0.282 & 0.242 & 0.399 & 0.323 & 0.340 & 0.066 & 0.110 \\ 
MedVInT \shortcite{zhang2023pmcvqa} & 0.443 & 0.349 & 0.288 & 0.246 & 0.404 & 0.326 & 0.377 & 0.148 & 0.212 \\ 
CT2Rep \shortcite{hamamci2024ct2rep} & 0.444 & 0.344 & 0.279 & 0.236 & 0.402 & 0.310 & 0.317 & 0.089 & 0.139 \\ 
M3D \shortcite{bai2024m3d} & 0.436 & 0.345 & 0.285 & 0.245 & 0.400 & 0.326 & 0.407 & 0.090 & 0.148 \\ 
RadFM \shortcite{wu2025towards} & 0.442 & 0.345 & 0.281 & 0.237 & 0.399 & 0.315 & 0.382 & 0.131 & 0.195 \\ 
Reg2RG \shortcite{chen2025large} & 0.473 & 0.365 & 0.296 & 0.249 & 0.441 & 0.367 & 0.423 & 0.181 & 0.253 \\ 
\ourmodel (\textbf{Ours}) & \textbf{0.482} & \textbf{0.375} & \textbf{0.305} & \textbf{0.257} & \textbf{0.456} & \textbf{0.383} & \textbf{0.495} & \textbf{0.335} & \textbf{0.399} \\
\bottomrule
\end{tabular}
\caption{Comparison of \ourmodel against state-of-the-art methods on RadGenome-ChestCT.} 
\label{tab:sota}
\end{table*}

\begin{table}[!h]
\small
\centering
\setlength{\tabcolsep}{4pt}
\begin{tabularx}{\linewidth}{X*{4}{c}}
\toprule
Method & BLEU-1 & BLEU-4 & METEOR & CE-F1 \\
\midrule
Resident-only & 0.469 & 0.246 & 0.435 & 0.219\\ 
SR-SA & 0.476 & 0.250 & 0.447 & 0.332\\
SR-MA & 0.475 & 0.251 & 0.454 & 0.352\\
MR-MA & 0.479 & 0.255 & 0.456 & 0.362\\
Ours & \textbf{0.482} &\textbf{ 0.257} & \textbf{0.456} & \textbf{0.399}\\
\bottomrule
\end{tabularx}
\caption{Ablation study of components in \ourmodel. SR-SA: Single Round Single Agent; SR-MA: Single Round Multi-Agent; MR-MA: Multi-Round Multi-Agent.}
\label{tab:ablation}
\end{table}

\begin{table}[ht]
\setlength{\tabcolsep}{4pt}
\small
\centering
\begin{tabularx}{\linewidth}{X*{4}{c}}
\toprule
LLM & BLEU-1 & BLEU-4 & METEOR & CE-F1 \\
\midrule
Resident-only & 0.469 & 0.246 & 0.435 & 0.219\\ 
GPT-4.1-mini & 0.480 & 0.255 & 0.454 & 0.393 \\
GPT-4.1 & 0.482 &0.257 & 0.456 & 0.399\\
GPT-4o & 0.479 & 0.255 & 0.454 & 0.392 \\
GPT-5 & 0.480 & 0.255 & 0.454 & 0.391 \\
\bottomrule
\end{tabularx}
\caption{Performance comparison of different LLMs.}
\label{tab:llm}
\end{table}

We evaluate \ourmodel on the RadGenome-ChestCT dataset~\cite{wu2025towards}, which contains 25,692 chest CT scans from 21,304 patients. Each CT scan is accompanied by a detailed radiology report authored by experienced radiologists.
We adhere to the official data split, using 24,128 scans for training and 1,564 scans for testing.
Statistics of the dataset are summarized in Appendix~\ref{appendix:dataset}.

We use GPT-4.1 and GPT-4o as the LLM backbones for \textit{Fellow} and \textit{Attending Agent}, respectively, with a temperature of 0 to ensure deterministic outputs.
The \textit{Resident} and \textit{Retrieval Agent} are implemented using the HuggingFace Transformers library~\cite{wolf2020transformers} and trained on a single NVIDIA H100 GPU.
The training process employs the AdamW optimizer with a learning rate of $1e-5$ and a batch size of 1.
We train \ourmodel for 10 epochs, which takes approximately 40 hours. 
Appendix~\ref{appendix:prompts} lists the template prompts for each agent.

We assess \ourmodel using standard natural language generation metrics that evaluate both lexical and semantic alignment with reference texts, including BLEU~\cite{papineni2002bleu}, ROUGE-L~\cite{lin2004rouge}, and METEOR~\cite{banerjee2005meteor}.
In addition, we assess clinical validity of the generated reports using the Clinical Efficacy (CE) score, which measures the accuracy of 18 predefined clinical abnormalities by computing precision, recall, and F1 scores with a pretrained RadBERT-RoBERTa-4m model~\citep{yan2022radbert}.

\section{Results and Discussions}

\textbf{Comparison with State-of-the-Art Methods.}
We benchmark \ourmodel against several state-of-the-art approaches for medical report generation (Table~\ref{tab:sota}). 
Across all evaluation metrics, \ourmodel consistently outperforms all baseline methods, demonstrating superior performance in generating high-quality and clinically accurate radiology reports.

\textbf{Ablation Study.}
To assess the contribution of each component in \ourmodel, we conduct an ablation study by removing or modifying key elements of the model.
Table~\ref{tab:ablation} indicates that each component significantly contributes to the overall performance, with the most notable drop observed when removing the consensus-driven Finalization.

\textbf{Sensitivity across LLMs.}
To investigate the impact of LLM size on report generation, we evaluate \ourmodel using different LLM backbones.
Table~\ref{tab:llm} shows all variants of \ourmodel consistently outperform baseline methods, with marginal performance improvements observed across advanced LLMs.

\begin{figure}[thbp] 
    \centering    
    \includegraphics[width=\linewidth]{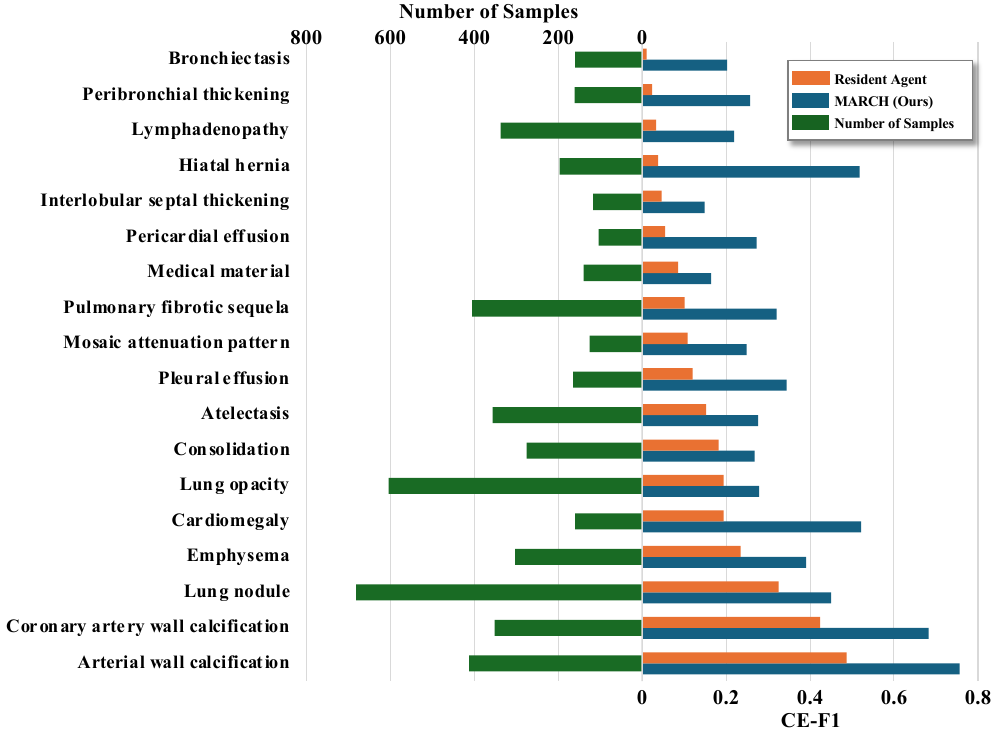}  
    \caption{Clinical efficacy across various abnormalities.}
    \label{fig:class_f1} 
\end{figure}
\textbf{Clinical Efficacy.}
Figure~\ref{fig:class_f1} presents the clinical efficacy of \ourmodel in F1-score across 18 abnormalities.
Compared with the baseline method, Resident Agent, \ourmodel demonstrates superior performance across most abnormalities, particularly in detecting minor abnormalities such as ``hiatal hernia'' and ``pericardial effusion'', indicating its enhanced ability to identify subtle clinical findings.
Detailed analysis is provided in Appendix~\ref{appendix:clinical_efficacy}.

\textbf{Case studies.} To further elucidate \ourmodel's capacity for generating interpretable and clinically grounded reports, we present a case study in Figure~\ref{fig:case_study} (Appendix~\ref{appendix:case_study}), which traces the hierarchical, multi-agent workflow intrinsic to our approach, with findings organized by anatomical region.

\textbf{Additional Results.}
We further provide additional experimental results, including the effectiveness of agent number (Appendix~\ref{appendix:ablation_agent_num}) and examples of generated reports (Appendix~\ref{appendix:examples}).
\section{Conclusions}
\label{sec:conclusion}
This paper presents \ourmodel, a consensus-driven, multi-agent framework to reduce cognitive errors in interpreting abnormal CT findings.  In contrast to prior work, \ourmodel coordinates agents for multi-scale 3D feature extraction, evidence-based retrieval augmented generation, and iterative consensus discourse.
Empirical evaluations demonstrate that \ourmodel significantly outperforms state-of-the-art models on both linguistic metrics and clinical fidelity, generates reports that reduce the risk of single-reader misinterpretation, and supports transparent, collaborative report generation. 

\section*{Limitations}
This method has demonstrated substantial improvements in medical image report generation. 
However, several limitations remain to be addressed in future work.
First, our current evaluation primarily utilizes the GPT series of large language models for multi-agent reasoning. Exploring the generalizability of this clinical hierarchy using diverse open-source or domain-specific medical LLMs is a critical next step.
Second, \ourmodel currently lacks a long-term memory mechanism, which limits its ability to incorporate longitudinal patient history or learn from past diagnostic errors across different cases.
Finally, while the framework emulates human clinical workflows, it operates as a fully autonomous system without a hybrid human-agent interface. Future iterations should investigate ``human-in-the-loop'' configurations where agents provide preliminary consensus reports for radiologist review and incorporate real-time clinical feedback to further bridge the gap between AI assistance and professional practice.
\section*{Acknowledgements}

This work was supported by the National Institutes of Health [grant numbers R01CA289249], the National Science Foundation (NSF) [grant numbers 2145640], the Patient-Centered Outcomes Research Institute (PCORI) [grant numbers ME-2023C3-35934],
and the Advanced Research Projects Agency for Health (ARPA-H) [grant name PARADIGM].
We gratefully acknowledge the support of NVIDIA Corporation and the NVIDIA AI Technology Center (NVAITC) UF program. 
\bibliography{ref}

@INPROCEEDINGS{dang2025multi-agent,
  title     = "Multi-Agent Collaboration via Evolving Orchestration",
  author    = "Dang, Yufan and Qian, Chen and Luo, Xueheng and Fan, Jingru and
               Xie, Zihao and Shi, Ruijie and Chen, Weize and Yang, Cheng and
               Che, Xiaoyin and Tian, Ye and Xiong, Xuantang and Han, Lei and
               Liu, Zhiyuan and Sun, Maosong",
  booktitle = "NeurIPS",
  month     =  "21~" # oct,
  year      =  2025,
  doi       = "10.48550/arXiv.2505.19591"
}

@INPROCEEDINGS{liao2025agentmaster,
  title     = "{AgentMaster}: A multi-agent conversational framework using {A2A}
               and {MCP} protocols for multimodal information retrieval and
               analysis",
  author    = "Liao, Callie C and Liao, Duoduo and Gadiraju, Sai Surya",
  booktitle = "Proceedings of the 2025 Conference on Empirical Methods in
               Natural Language Processing: System Demonstrations",
  publisher = "Association for Computational Linguistics",
  address   = "Stroudsburg, PA, USA",
  pages     = "52--72",
  month     =  nov,
  year      =  2025,
  doi       = "10.18653/v1/2025.emnlp-demos.5"
}

@INPROCEEDINGS{du2025multi-agent,
  title     = "Multi-Agent Collaboration via Cross-Team Orchestration",
  author    = "Du, Zhuoyun and Qian, Chen and Liu, Wei and Xie, Zihao and Wang,
               Yifei and Qiu, Rennai and Dang, Yufan and Chen, Weize and Yang,
               Cheng and Tian, Ye and Xiong, Xuantang and Han, Lei",
  booktitle = "Findings of the Association for Computational Linguistics: ACL
               2025",
  publisher = "Association for Computational Linguistics",
  address   = "Stroudsburg, PA, USA",
  pages     = "10386--10406",
  year      =  2025,
  doi       = "10.18653/v1/2025.findings-acl.541"
}

@ARTICLE{waite2017interpretive,
  title     = "Interpretive error in radiology",
  author    = "Waite, Stephen and Scott, Jinel and Gale, Brian and Fuchs, Travis
               and Kolla, Srinivas and Reede, Deborah",
  journal   = "AJR Am. J. Roentgenol.",
  publisher = "American Roentgen Ray Society",
  volume    =  208,
  number    =  4,
  pages     = "739--749",
  month     =  apr,
  year      =  2017,
  doi       = "10.2214/AJR.16.16963",
  pmid      =  28026210,
  issn      = "0361-803X,1546-3141"
}

@ARTICLE{seah2021effect,
  title     = "Effect of a comprehensive deep-learning model on the accuracy of
               chest x-ray interpretation by radiologists: a retrospective,
               multireader multicase study",
  author    = "Seah, Jarrel C Y and Tang, Cyril H M and Buchlak, Quinlan D and
               Holt, Xavier G and Wardman, Jeffrey B and Aimoldin, Anuar and
               Esmaili, Nazanin and Ahmad, Hassan and Pham, Hung and Lambert,
               John F and Hachey, Ben and Hogg, Stephen J F and Johnston,
               Benjamin P and Bennett, Christine and Oakden-Rayner, Luke and
               Brotchie, Peter and Jones, Catherine M",
  journal   = "Lancet Digit. Health",
  publisher = "Elsevier BV",
  volume    =  3,
  number    =  8,
  pages     = "e496--e506",
  month     =  aug,
  year      =  2021,
  doi       = "10.1016/S2589-7500(21)00106-0",
  pmid      =  34219054,
  issn      = "2589-7500"
}

@article{touvron2023llama,
  title         = "Llama 2: Open Foundation and Fine-Tuned Chat Models",
  author        = "Touvron, Hugo and Martin, Louis and Stone, Kevin and Albert,
                   Peter and Almahairi, Amjad and Babaei, Yasmine and Bashlykov,
                   Nikolay and Batra, Soumya and Bhargava, Prajjwal and Bhosale,
                   Shruti and Bikel, Dan and Blecher, Lukas and Ferrer, Cristian
                   Canton and Chen, Moya and Cucurull, Guillem and Esiobu, David
                   and Fernandes, Jude and Fu, Jeremy and Fu, Wenyin and Fuller,
                   Brian and Gao, Cynthia and Goswami, Vedanuj and Goyal, Naman
                   and Hartshorn, Anthony and Hosseini, Saghar and Hou, Rui and
                   Inan, Hakan and Kardas, Marcin and Kerkez, Viktor and Khabsa,
                   Madian and Kloumann, Isabel and Korenev, Artem and Koura,
                   Punit Singh and Lachaux, Marie-Anne and Lavril, Thibaut and
                   Lee, Jenya and Liskovich, Diana and Lu, Yinghai and Mao,
                   Yuning and Martinet, Xavier and Mihaylov, Todor and Mishra,
                   Pushkar and Molybog, Igor and Nie, Yixin and Poulton, Andrew
                   and Reizenstein, Jeremy and Rungta, Rashi and Saladi, Kalyan
                   and Schelten, Alan and Silva, Ruan and Smith, Eric Michael
                   and Subramanian, Ranjan and Tan, Xiaoqing Ellen and Tang,
                   Binh and Taylor, Ross and Williams, Adina and Kuan, Jian
                   Xiang and Xu, Puxin and Yan, Zheng and Zarov, Iliyan and
                   Zhang, Yuchen and Fan, Angela and Kambadur, Melanie and
                   Narang, Sharan and Rodriguez, Aurelien and Stojnic, Robert
                   and Edunov, Sergey and Scialom, Thomas",
  journal       = "arXiv [cs.CL]",
  month         =  "18~" # jul,
  year          =  2023,
  archivePrefix = "arXiv",
  primaryClass  = "cs.CL",
  eprint        = "2307.09288"
}

@article{wang2023r2gengpt,
  title     = "{R2GenGPT}: Radiology Report Generation with frozen {LLMs}",
  author    = "Wang, Zhanyu and Liu, Lingqiao and Wang, Lei and Zhou, Luping",
  journal   = "Meta-Radiology",
  publisher = "Elsevier BV",
  volume    =  1,
  number    =  3,
  pages     =  100033,
  month     =  "1~" # nov,
  year      =  2023,
  doi       = "10.1016/j.metrad.2023.100033",
  issn      = "2950-1628"
}

@article{zhang2023pmcvqa,
  title     = "Development of a large-scale medical visual question-answering
               dataset",
  author    = "Zhang, Xiaoman and Wu, Chaoyi and Zhao, Ziheng and Lin, Weixiong
               and Zhang, Ya and Wang, Yanfeng and Xie, Weidi",
  journal   = "Communications Medicine volume",
  volume    =  4,
  number    =  1,
  pages     =  277,
  month     =  "21~" # dec,
  year      =  2024,
  doi       = "10.1038/s43856-024-00709-2",
  pmc       = "PMC11663219",
  pmid      =  39709495,
  issn      = "2730-664X"
}

@article{wu2025towards,
  title     = "Towards generalist foundation model for radiology by leveraging
               web-scale {2D\&3D} medical data",
  author    = "Wu, Chaoyi and Zhang, Xiaoman and Zhang, Ya and Hui, Hui and
               Wang, Yanfeng and Xie, Weidi",
  journal   = "Nature Communications",
  publisher = "Springer Science and Business Media LLC",
  volume    =  16,
  number    =  1,
  pages     =  7866,
  month     =  "23~" # aug,
  year      =  2025,
  doi       = "10.1038/s41467-025-62385-7",
  pmc       = "PMC12375113",
  pmid      =  40849424,
  issn      = "2041-1723"
}

@INCOLLECTION{hamamci2024ct2rep,
  title     = "{CT2Rep}: Automated radiology report generation for {3D} medical
               imaging",
  author    = "Hamamci, Ibrahim Ethem and Er, Sezgin and Menze, Bjoern",
  booktitle = "Medical Image Computing and Computer Assisted Intervention",
  publisher = "Springer Nature Switzerland",
  address   = "Cham",
  pages     = "476--486",
  series    = "Lecture notes in computer science",
  year      =  2024,
  doi       = "10.1007/978-3-031-72390-2\_45",
  issn      = "1611-3349,0302-9743"
}

@article{bai2024m3d,
  title         = "{M3D}: Advancing {3D} medical image analysis with multi-modal large language models",
  author        = "Bai, Fan and Du, Yuxin and Huang, Tiejun and Meng, Max Q-H
                   and Zhao, Bo",
  journal       = "arXiv [cs.CV]",
  month         =  "31~" # mar,
  year          =  2024,
  archivePrefix = "arXiv",
  primaryClass  = "cs.CV",
  eprint        = "2404.00578"
}

@article{chen2025large,
  title     = "Large language model with region-guided referring and grounding
               for {CT} report generation",
  author    = "Chen, Zhixuan and Bie, Yequan and Jin, Haibo and Chen, Hao",
  journal   = "IEEE transactions on Medical Imaging",
  publisher = "Institute of Electrical and Electronics Engineers (IEEE)",
  volume    =  44,
  number    =  8,
  pages     = "3139--3150",
  month     =  aug,
  year      =  2025,
  doi       = "10.1109/TMI.2025.3559923",
  pmid      =  40215158,
  issn      = "1558-254X,0278-0062"
}

@ARTICLE{zhao2025large-vocabulary,
  title     = "Large-vocabulary segmentation for medical images with text
               prompts",
  author    = "Zhao, Ziheng and Zhang, Yao and Wu, Chaoyi and Zhang, Xiaoman and
               Zhou, Xiao and Zhang, Ya and Wang, Yanfeng and Xie, Weidi",
  journal   = "NPJ Digital Medicine",
  publisher = "Springer Science and Business Media LLC",
  volume    =  8,
  number    =  1,
  pages     =  566,
  month     =  "2~" # sep,
  year      =  2025,
  doi       = "10.1038/s41746-025-01964-w",
  pmc       = "PMC12405521",
  pmid      =  40897901,
  issn      = "2398-6352"
}

@inproceedings{hu2022lora,
  title     = "{LoRA}: Low-Rank Adaptation of Large Language Models",
  author    = "Hu, Edward J and Wallis, Phillip and Allen-Zhu, Zeyuan and Li,
               Yuanzhi and Wang, Shean and Wang, Lu and Chen, Weizhu and
               {Others}",
  booktitle = "International Conference on Learning Representations",
  pages     = "1--26",
  year      =  2021
}

@inproceedings{papineni2002bleu,
  title     = "{BLEU}: a method for automatic evaluation of machine translation",
  author    = "Papineni, Kishore and Roukos, Salim and Ward, Todd and Zhu,
               Wei-Jing",
  booktitle = "Proceedings of the 40th Annual Meeting on Association for
               Computational Linguistics",
  publisher = "Association for Computational Linguistics",
  address   = "USA",
  pages     = "311--318",
  series    = "ACL '02",
  month     =  "6~" # jul,
  year      =  2002,
  doi       = "10.3115/1073083.1073135"
}

@inproceedings{lin2004rouge,
  title     = "{ROUGE}: a package for automatic evaluation of summaries",
  author    = "Lin, Chin-Yew",
  booktitle = "Text summarization branches out: Proceedings of the ACL-04
               workshop",
  volume    =  8,
  pages     = "1--8",
  year      =  2004
}

@inproceedings{banerjee2005meteor,
  title     = "{METEOR}: an automatic metric for {MT} evaluation with improved
               correlation with human judgments",
  author    = "Banerjee, Satanjeev and Lavie, Alon",
  booktitle = "Proceedings of the ACL Workshop on Intrinsic and Extrinsic
               Evaluation Measures for Machine Translation and/or Summarization",
  pages     = "65--72",
  year      =  2005
}

@article{yan2022radbert,
  title     = "{RadBERT}: Adapting transformer-based language models to
               radiology",
  author    = "Yan, An and McAuley, Julian and Lu, Xing and Du, Jiang and Chang,
               Eric Y and Gentili, Amilcare and Hsu, Chun-Nan",
  journal   = "Radiol. Artif. Intell.",
  publisher = "Radiological Society of North America",
  volume    =  4,
  number    =  4,
  pages     = "e210258",
  month     =  jul,
  year      =  2022,
  doi       = "10.1148/ryai.210258",
  pmc       = "PMC9344353",
  pmid      =  35923376,
  issn      = "2638-6100"
}

@misc{openai2024gpt4technicalreport,
      title={{GPT}-4 Technical Report}, 
      author={OpenAI and Josh Achiam and Steven Adler and Sandhini Agarwal and Lama Ahmad and Ilge Akkaya and Florencia Leoni Aleman and Diogo Almeida and Janko Altenschmidt and Sam Altman and Shyamal Anadkat and Red Avila and Igor Babuschkin and Suchir Balaji and Valerie Balcom and Paul Baltescu and Haiming Bao and Mohammad Bavarian and Jeff Belgum and Irwan Bello and Jake Berdine and Gabriel Bernadett-Shapiro and Christopher Berner and Lenny Bogdonoff and Oleg Boiko and Madelaine Boyd and Anna-Luisa Brakman and Greg Brockman and Tim Brooks and Miles Brundage and Kevin Button and Trevor Cai and Rosie Campbell and Andrew Cann and Brittany Carey and Chelsea Carlson and Rory Carmichael and Brooke Chan and Che Chang and Fotis Chantzis and Derek Chen and Sully Chen and Ruby Chen and Jason Chen and Mark Chen and Ben Chess and Chester Cho and Casey Chu and Hyung Won Chung and Dave Cummings and Jeremiah Currier and Yunxing Dai and Cory Decareaux and Thomas Degry and Noah Deutsch and Damien Deville and Arka Dhar and David Dohan and Steve Dowling and Sheila Dunning and Adrien Ecoffet and Atty Eleti and Tyna Eloundou and David Farhi and Liam Fedus and Niko Felix and Simón Posada Fishman and Juston Forte and Isabella Fulford and Leo Gao and Elie Georges and Christian Gibson and Vik Goel and Tarun Gogineni and Gabriel Goh and Rapha Gontijo-Lopes and Jonathan Gordon and Morgan Grafstein and Scott Gray and Ryan Greene and Joshua Gross and Shixiang Shane Gu and Yufei Guo and Chris Hallacy and Jesse Han and Jeff Harris and Yuchen He and Mike Heaton and Johannes Heidecke and Chris Hesse and Alan Hickey and Wade Hickey and Peter Hoeschele and Brandon Houghton and Kenny Hsu and Shengli Hu and Xin Hu and Joost Huizinga and Shantanu Jain and Shawn Jain and Joanne Jang and Angela Jiang and Roger Jiang and Haozhun Jin and Denny Jin and Shino Jomoto and Billie Jonn and Heewoo Jun and Tomer Kaftan and Łukasz Kaiser and Ali Kamali and Ingmar Kanitscheider and Nitish Shirish Keskar and Tabarak Khan and Logan Kilpatrick and Jong Wook Kim and Christina Kim and Yongjik Kim and Jan Hendrik Kirchner and Jamie Kiros and Matt Knight and Daniel Kokotajlo and Łukasz Kondraciuk and Andrew Kondrich and Aris Konstantinidis and Kyle Kosic and Gretchen Krueger and Vishal Kuo and Michael Lampe and Ikai Lan and Teddy Lee and Jan Leike and Jade Leung and Daniel Levy and Chak Ming Li and Rachel Lim and Molly Lin and Stephanie Lin and Mateusz Litwin and Theresa Lopez and Ryan Lowe and Patricia Lue and Anna Makanju and Kim Malfacini and Sam Manning and Todor Markov and Yaniv Markovski and Bianca Martin and Katie Mayer and Andrew Mayne and Bob McGrew and Scott Mayer McKinney and Christine McLeavey and Paul McMillan and Jake McNeil and David Medina and Aalok Mehta and Jacob Menick and Luke Metz and Andrey Mishchenko and Pamela Mishkin and Vinnie Monaco and Evan Morikawa and Daniel Mossing and Tong Mu and Mira Murati and Oleg Murk and David Mély and Ashvin Nair and Reiichiro Nakano and Rajeev Nayak and Arvind Neelakantan and Richard Ngo and Hyeonwoo Noh and Long Ouyang and Cullen O'Keefe and Jakub Pachocki and Alex Paino and Joe Palermo and Ashley Pantuliano and Giambattista Parascandolo and Joel Parish and Emy Parparita and Alex Passos and Mikhail Pavlov and Andrew Peng and Adam Perelman and Filipe de Avila Belbute Peres and Michael Petrov and Henrique Ponde de Oliveira Pinto and Michael and Pokorny and Michelle Pokrass and Vitchyr H. Pong and Tolly Powell and Alethea Power and Boris Power and Elizabeth Proehl and Raul Puri and Alec Radford and Jack Rae and Aditya Ramesh and Cameron Raymond and Francis Real and Kendra Rimbach and Carl Ross and Bob Rotsted and Henri Roussez and Nick Ryder and Mario Saltarelli and Ted Sanders and Shibani Santurkar and Girish Sastry and Heather Schmidt and David Schnurr and John Schulman and Daniel Selsam and Kyla Sheppard and Toki Sherbakov and Jessica Shieh and Sarah Shoker and Pranav Shyam and Szymon Sidor and Eric Sigler and Maddie Simens and Jordan Sitkin and Katarina Slama and Ian Sohl and Benjamin Sokolowsky and Yang Song and Natalie Staudacher and Felipe Petroski Such and Natalie Summers and Ilya Sutskever and Jie Tang and Nikolas Tezak and Madeleine B. Thompson and Phil Tillet and Amin Tootoonchian and Elizabeth Tseng and Preston Tuggle and Nick Turley and Jerry Tworek and Juan Felipe Cerón Uribe and Andrea Vallone and Arun Vijayvergiya and Chelsea Voss and Carroll Wainwright and Justin Jay Wang and Alvin Wang and Ben Wang and Jonathan Ward and Jason Wei and CJ Weinmann and Akila Welihinda and Peter Welinder and Jiayi Weng and Lilian Weng and Matt Wiethoff and Dave Willner and Clemens Winter and Samuel Wolrich and Hannah Wong and Lauren Workman and Sherwin Wu and Jeff Wu and Michael Wu and Kai Xiao and Tao Xu and Sarah Yoo and Kevin Yu and Qiming Yuan and Wojciech Zaremba and Rowan Zellers and Chong Zhang and Marvin Zhang and Shengjia Zhao and Tianhao Zheng and Juntang Zhuang and William Zhuk and Barret Zoph},
      year={2024},
      eprint={2303.08774},
      archivePrefix={arXiv},
      primaryClass={cs.CL},
      url={https://arxiv.org/abs/2303.08774}, 
}

@article{khan2025comprehensive,
  title={A comprehensive survey of foundation models in medicine},
  author={Khan, Wasif and Leem, Seowung and See, Kyle B and Wong, Joshua K and Zhang, Shaoting and Fang, Ruogu},
  journal={IEEE Reviews in Biomedical Engineering},
  year={2025},
  publisher={IEEE}
}

@article{plaat2025agentic,
  title={Agentic large language models, a survey},
  author={Plaat, Aske and van Duijn, Max and van Stein, Niki and Preuss, Mike and van der Putten, Peter and Batenburg, Kees Joost},
  journal={arXiv preprint arXiv:2503.23037},
  year={2025}
}

@article{singhal2025toward,
  title={Toward expert-level medical question answering with large language models},
  author={Singhal, Karan and Tu, Tao and Gottweis, Juraj and Sayres, Rory and Wulczyn, Ellery and Amin, Mohamed and Hou, Le and Clark, Kevin and Pfohl, Stephen R and Cole-Lewis, Heather and others},
  journal={Nature Medicine},
  volume={31},
  number={3},
  pages={943--950},
  year={2025},
  publisher={Nature Publishing Group US New York}
}

@article{goh2024large,
  title={Large language model influence on diagnostic reasoning: a randomized clinical trial},
  author={Goh, Ethan and Gallo, Robert and Hom, Jason and Strong, Eric and Weng, Yingjie and Kerman, Hannah and Cool, Jos{\'e}phine A and Kanjee, Zahir and Parsons, Andrew S and Ahuja, Neera and others},
  journal={JAMA network open},
  volume={7},
  number={10},
  pages={e2440969--e2440969},
  year={2024},
  publisher={American Medical Association}
}

@article{yamamoto2024enhancing,
  title={Enhancing medical interview skills through AI-Simulated patient interactions: nonrandomized controlled trial},
  author={Yamamoto, Akira and Koda, Masahide and Ogawa, Hiroko and Miyoshi, Tomoko and Maeda, Yoshinobu and Otsuka, Fumio and Ino, Hideo and others},
  journal={JMIR medical education},
  volume={10},
  number={1},
  pages={e58753},
  year={2024},
  publisher={JMIR Publications Inc., Toronto, Canada}
}

@article{wang2025accelerating,
  title={Accelerating clinical evidence synthesis with large language models},
  author={Wang, Zifeng and Cao, Lang and Danek, Benjamin and Jin, Qiao and Lu, Zhiyong and Sun, Jimeng},
  journal={npj Digital Medicine},
  volume={8},
  number={1},
  pages={509},
  year={2025},
  publisher={Nature Publishing Group UK London}
}

@article{zhong2025vision,
  title={Vision-language model for report generation and outcome prediction in {CT} pulmonary angiogram},
  author={Zhong, Zhusi and Wang, Yuli and Wu, Jing and Hsu, Wen-Chi and Somasundaram, Vin and Bi, Lulu and Kulkarni, Shreyas and Ma, Zhuoqi and Collins, Scott and Baird, Grayson and others},
  journal={NPJ Digital Medicine},
  volume={8},
  number={1},
  pages={432},
  year={2025},
  publisher={Nature Publishing Group UK London}
}

@inproceedings{liu2025argus,
  title={Argus: benchmarking and enhancing vision-language models for 3D radiology report generation},
  author={Liu, Che and Wan, Zhongwei and Wang, Yuqi and Shen, Hui and Wang, Haozhe and Zheng, Kangyu and Zhang, Mi and Arcucci, Rossella},
  booktitle={Findings of the Association for Computational Linguistics: ACL 2025},
  pages={16448--16460},
  year={2025}
}

@inproceedings{jiang2025comt,
  title={Comt: Chain-of-medical-thought reduces hallucination in medical report generation},
  author={Jiang, Yue and Chen, Jiawei and Yang, Dingkang and Li, Mingcheng and Wang, Shunli and Wu, Tong and Li, Ke and Zhang, Lihua},
  booktitle={ICASSP 2025-2025 IEEE International Conference on Acoustics, Speech and Signal Processing (ICASSP)},
  pages={1--5},
  year={2025},
  organization={IEEE}
}

@article{johnson2019mimic,
  title={{MIMIC-CXR-JPG}, a large publicly available database of labeled chest radiographs},
  author={Johnson, Alistair EW and Pollard, Tom J and Greenbaum, Nathaniel R and Lungren, Matthew P and Deng, Chih-ying and Peng, Yifan and Lu, Zhiyong and Mark, Roger G and Berkowitz, Seth J and Horng, Steven},
  journal={arXiv preprint arXiv:1901.07042},
  year={2019}
}

@article{yi2025multimodal,
  title={A Multimodal Multi-Agent Framework for Radiology Report Generation},
  author={Yi, Ziruo and Xiao, Ting and Albert, Mark V},
  journal={arXiv preprint arXiv:2505.09787},
  year={2025}
}

@inproceedings{fan2025ai,
  title={{AI} hospital: Benchmarking large language models in a multi-agent medical interaction simulator},
  author={Fan, Zhihao and Wei, Lai and Tang, Jialong and Chen, Wei and Siyuan, Wang and Wei, Zhongyu and Huang, Fei},
  booktitle={Proceedings of the 31st International Conference on Computational Linguistics},
  pages={10183--10213},
  year={2025}
}

@inproceedings{chen2025self,
  title={A Self-Evolving Framework for Multi-Agent Medical Consultation Based on Large Language Models},
  author={Chen, Kai and Qi, Ji and Huo, Jing and Tian, Pinzhuo and Meng, Fanyu and Yang, Xi and Gao, Yang},
  booktitle={ICASSP 2025-2025 IEEE International Conference on Acoustics, Speech and Signal Processing (ICASSP)},
  pages={1--5},
  year={2025},
  organization={IEEE}
}

@article{kim2024mdagents,
  title={Mdagents: An adaptive collaboration of llms for medical decision-making},
  author={Kim, Yubin and Park, Chanwoo and Jeong, Hyewon and Chan, Yik S and Xu, Xuhai and McDuff, Daniel and Lee, Hyeonhoon and Ghassemi, Marzyeh and Breazeal, Cynthia and Park, Hae W},
  journal={Advances in Neural Information Processing Systems},
  volume={37},
  pages={79410--79452},
  year={2024}
}

@inproceedings{li2024mmedagent,
  title={{MMedAgent}: Learning to Use Medical Tools with Multi-modal Agent},
  author={Li, Binxu and Yan, Tiankai and Pan, Yuanting and Luo, Jie and Ji, Ruiyang and Ding, Jiayuan and Xu, Zhe and Liu, Shilong and Dong, Haoyu and Lin, Zihao and others},
  booktitle={Findings of the Association for Computational Linguistics: EMNLP 2024},
  pages={8745--8760},
  year={2024}
}

@article{sloan2024automated,
  title={Automated radiology report generation: A review of recent advances},
  author={Sloan, Phillip and Clatworthy, Philip and Simpson, Edwin and Mirmehdi, Majid},
  journal={IEEE Reviews in Biomedical Engineering},
  volume={18},
  pages={368--387},
  year={2024},
  publisher={IEEE}
}

@article{ma2025fully,
  title={A fully open {AI} foundation model applied to chest radiography},
  author={Ma, DongAo and Pang, Jiaxuan and Gotway, Michael B and Liang, Jianming},
  journal={Nature},
  pages={1--11},
  year={2025},
  publisher={Nature Publishing Group UK London}
}

@inproceedings{liu2024context,
  title={In-context learning for zero-shot medical report generation},
  author={Liu, Rui and Li, Mingjie and Zhao, Shen and Chen, Ling and Chang, Xiaojun and Yao, Lina},
  booktitle={Proceedings of the 32nd ACM international conference on multimedia},
  pages={8721--8730},
  year={2024}
}

@inproceedings{liu2024bootstrapping,
  title={Bootstrapping large language models for radiology report generation},
  author={Liu, Chang and Tian, Yuanhe and Chen, Weidong and Song, Yan and Zhang, Yongdong},
  booktitle={Proceedings of the AAAI Conference on Artificial Intelligence},
  volume={38},
  pages={18635--18643},
  year={2024}
}

@article{moor2023foundation,
  title={Foundation models for generalist medical artificial intelligence},
  author={Moor, Michael and Banerjee, Oishi and Abad, Zahra Shakeri Hossein and Krumholz, Harlan M and Leskovec, Jure and Topol, Eric J and Rajpurkar, Pranav},
  journal={Nature},
  volume={616},
  number={7956},
  pages={259--265},
  year={2023},
  publisher={Nature Publishing Group UK London}
}

@inproceedings{wang2023metransformer,
  title={Metransformer: Radiology report generation by transformer with multiple learnable expert tokens},
  author={Wang, Zhanyu and Liu, Lingqiao and Wang, Lei and Zhou, Luping},
  booktitle={Proceedings of the IEEE/CVF Conference on Computer Vision and Pattern Recognition},
  pages={11558--11567},
  year={2023}
}

@inproceedings{zhu2025can,
  title={Can we trust AI doctors? a survey of medical hallucination in large language and large vision-language models},
  author={Zhu, Zhihong and Zhang, Yunyan and Zhuang, Xianwei and Zhang, Fan and Wan, Zhongwei and Chen, Yuyan and QingqingLong, QingqingLong and Zheng, Yefeng and Wu, Xian},
  booktitle={Findings of the Association for Computational Linguistics: ACL 2025},
  pages={6748--6769},
  year={2025}
}

@article{hill2017can,
  title={How can surgeons facilitate resident intraoperative decision-making?},
  author={Hill, Katherine A and Dasari, Mohini and Littleton, Eliza B and Hamad, Giselle G},
  journal={The American Journal of Surgery},
  volume={214},
  number={4},
  pages={583--588},
  year={2017},
  publisher={Elsevier}
}

@article{goold2006ethics,
  title={Ethics and professionalism: what does a resident need to learn?},
  author={Goold, Susan Dorr and Stern, David T},
  journal={The American Journal of Bioethics},
  volume={6},
  number={4},
  pages={9--17},
  year={2006},
  publisher={Taylor \& Francis}
}

@article{bolin2006strategies,
  title={Strategies for incorporating professional ethics education in graduate medical programs},
  author={Bolin, Jane Nelson},
  journal={The American Journal of Bioethics},
  volume={6},
  number={4},
  pages={35--36},
  year={2006},
  publisher={Taylor \& Francis}
}

@inproceedings{wolf2020transformers,
  title={Transformers: State-of-the-art natural language processing},
  author={Wolf, Thomas and Debut, Lysandre and Sanh, Victor and Chaumond, Julien and Delangue, Clement and Moi, Anthony and Cistac, Pierric and Rault, Tim and Louf, Remi and Funtowicz, Morgan and others},
  booktitle={Proceedings of the 2020 conference on empirical methods in natural language processing: system demonstrations},
  pages={38--45},
  year={2020}
}
\clearpage
\appendix
\centerline{\Large\textbf{Appendix}}
\section{Example Prompt Template}
\label{appendix:prompts}
Here we present an example of the prompt template used for the Resident Agent $\mathcal{A}_{\text{res}}$ in the Initial Report Drafting, Retrieval-Augmented Revision, and Consensus-Driven Finalization phases of \ourmodel.

\begin{prompt}[Resident Agent ($\mathcal{A}_{\text{res}}$) for Initial Report Drafting]
The global information is provided as the context: \\ 
<image\_1> <image\_2> ... <image\_n>. \\
The region 1 is \\
<region\_1> <region\_2> ... <region\_n>. \\
The region 2 is \\
<region\_(n+1)> <region\_(n+2)> ... <region\_2n>. \\
...\\
The region 10 is \\
<region\_(9n+1)> <region\_(9n+2)> ... <region\_10n>. \\

Given the provided global and regional information from this CT scan, please generate a comprehensive medical report for each region. First, identify the anatomical area corresponding to each region, then provide detailed information about these anatomical structures and any abnormalities that are essential. You can refer to the global information as the context and take it as a supplement. 
\end{prompt}

\begin{prompt}[Fellow Agent ($\mathcal{A}_{\text{fel}}$) for Consensus Synthesis]
You are an experienced medical doctor. Your task is to review and analyze a patient's medical report based on the retrieved relevant medical reports from the database. You need to carefully compare the initial medical report with the retrieved relevant medical reports, identify any discrepancies or inconsistencies, and make necessary modifications to ensure the accuracy and coherence of the final medical report. \\

Please follow these steps to complete your task: \\
1. Carefully read and understand both the initial medical report and the retrieved relevant medical reports. \\
2. Identify any discrepancies, conflicts, or inconsistencies between the two reports. \\
3. Make necessary modifications to the initial medical report to resolve any identified issues, ensuring that the final report is clinically accurate and coherent. \\
4. If no changes are necessary, output the original initial medical report as is. \\

Here is the initial medical report:
<init\_report> \\

Here are the retrieved relevant medical reports from the database:
<retrieved\_report> \\

Here is an example of the format you should output:
\{"report": "The region 0 is abdomen: ... The region 1 is bone: ... The region 2 is breast: ..."\}.

Respond in JSON format without any additional content.
\end{prompt}

\begin{prompt}[Attending Agent ($\mathcal{A}_{\text{att}}$) for Consensus Synthesis]
You are an authoritative expert in the medical field. You are organizing a collaborative consultation. Now several doctors have made patient's medical report based on the retrieved relevant medical reports from the database. Your task is to analyze the rationality of each doctor's opinion, summarize the opinions to obtain a synthesized report for the patient and give your final medical report. \\

First, please read the patient's initial medical report, as follows:
<init\_report> \\

Then, all doctors make a revised medical report based on the retrieved relevant medical reports from the database.
The following are their opinions:
<doctor\_info> 
You need to read all doctors' opinions carefully and analyze whether their opinions make sense. \\

Next, please write a synthesized report including the following: \\
1. Final medical report after your analysis. \\
2. A List of supporting evidence represented as a string. Please output detailed content from some of the evidence provided by the doctors or some analysis results of the doctors. Do not just list the doctor's name. \\

Please follow the guidelines below to complete your task: \\
1. Carefully read and understand both the initial medical report and the retrieved relevant medical reports. \\
2. Identify any discrepancies, conflicts, or inconsistencies between the two reports. \\
3. Make necessary modifications to the initial medical report to resolve any identified issues, ensuring that the final report is clinically accurate and coherent. \\
4. If no changes are necessary, output the original initial medical report as is. \\

Here is an example of the format you should output:
\{"report": "The region 0 is abdomen: ... The region 1 is bone: ... The region 2 is breast: ...", "reasons": ["...", "..."]\}.

Respond in JSON format without any additional content.
\end{prompt}

\begin{prompt}[Fellow Agent for Iterative Refinement]
You are an experienced radiology fellow participating in a consultation with several other medical doctors for a patient. The attending radiologist of this consultation has generated a revised report based on all doctors' analysis of the patient. Please provide your viewpoint on his opinion. \\

Here is the relevant medical knowledge:
{{retrieved\_report}} \\

Here is the initial medical report:
{{init\_report}} \\

Here is your last analysis of the patient, which is not completely reasonable, and you may need to adjust it based on the attending radiologist's opinion:
{{fellow\_report}} \\

Here is the revised report generated by the attending radiologist:
{{attending\_report}} \\

Here are the reasons provided by the attending radiologist:
{{attending\_reason}} \\

You need to consider the attending radiologist's opinion carefully and provide your opinions on the revised report generated by the attending radiologist. Please output your opinions including the following content: \\
1. Your viewpoint on the opinion of the attending radiologist, i.e., respond with "agree" or "disagree". \\
2. The confidence score of your opinion, respond with an integer between 1 and 3. The meaning of the confidence score is as follows: \\
    3 for High - You are an expert in the subject area and have extensive knowledge in the medical domain. You are highly confident in your ability to provide an accurate and thorough assessment. Your evaluation is based on deep expertise and a comprehensive understanding of the work. \\
    2 for Moderate - You have a good understanding of the subject area and is familiar with the medical domain. You feel confident in your ability to accurately assess the quality and significance of the work. Your evaluation is based on a solid grasp of the content and context. \\
    1 for Low - You have some knowledge of the subject area and is somewhat familiar with the medical domain. You understand the main points but may lack depth in certain areas. You are reasonably confident in your assessment but acknowledges some limitations in your expertise. \\
3. The reason for your opinion. If you change your opinion, for example, you agree with the attending radiologist's opinion which is different from your last analysis, please respond that you have changed in your response and provide detailed reasons for the change. You need to point out the parts where you got the wrong conclusion or the important parts you ignored in your last analysis, and the new key features that you think are important and the impact of these features on the patient's clinical abnormalities.
4. The evidence you use to support your opinion. Please choose from the relevant medical knowledge I provide as your evidence, and must output important content from the evidence. \\

Here are examples of the format you should output: \\
{"answer": "agree", "confidence": 3, "reason": "The reason for your opinion.", "evidences": ["Evidence 1 ...", "Evidence 2 ..."]}, \\
{"answer": "disagree", "confidence": 1, "reason": "The reason for your opinion.", "evidences": ["Evidence 1 ...", "Evidence 2 ..."]} 

Respond in JSON format without any additional content.
\end{prompt}

\begin{prompt}[Attending Agent ($\mathcal{A}_{\text{att}}$) for Iterative Refinement]
You are an authoritative expert in the medical field. You are organizing a collaborative consultation. Now several doctors have made analysis and judgments on a your previous report. Your task is to judge whether everyone has reached a consensus on the medical report  based on the analysis statements of each doctor and then analyze the rationality of each doctor's opinion and give your final medical report. \\

In the previous discussion, you took into account the opinions of all the doctors and obtained a report about the patient, which is listed as follows:
<current\_report> \\

In response to the patient's synthesized report, several doctors put forward their own opinions and reasons.
In each doctor's statement, they first express whether they agree with your statement in the previous synthesized report and give the confidence level on their own judgment. Then, they further elaborate on their views by stating reasons and listing relevant references.
The following are their opinions:
<fellow\_info>\\

Now, you need to judge whether the next round of discussion is needed based on each doctor's statement. Considering the following four cases: \\
1. If all doctors agree with the previous synthesized report, there is no need to continue the discussion. \\
2. If some doctors disagree with the previous report, but they are not confident in their judgment and have not listed convincing evidence, there is no need to continue the discussion. \\
3. If some doctors strongly oppose the previous report and you think their evidence is worth discussing, please continue the discussion. \\
4. If most doctors disagree with the previous report, please continue the discussion. \\

If you think the discussion should continue, you need to analyze the rationality of each doctor's opinions and summarize the opinions you think are reasonable to obtain a synthesized report for the patient. You should follow these cases: \\
1. If a doctor expresses strong opposition to your previous report, you need to focus on his\/her reasons and arguments and think carefully about whether you need to reconsider the diagnosis of the patient and modify your synthesized report accordingly. \\
2. If a doctor expresses opposition but also has some doubts about his\/her own opinion, you need to consider his\/her opinion, but you can stick to your original opinion.
3. If a doctor expresses agreement, then you do not need to modify your original synthesized report based on his\/her opinion.

Please output the following four contents:
1. Whether to continue the discussion or not. Please respond with `Yes` or `No`. \\
2. Your revised report. \\
3. Your reasons for revision. The format of the reason for revision is: which doctor's opinion or relevant literature you refer to, and which original opinions you modify. Please output detailed content, don't just list the doctor's name. \\
4. The instructions for each doctor to follow in the next round of discussion. The instruction should be specific and actionable, guiding each doctor on how to adjust their analysis or what aspects to focus on based on the previous round's discussion. \\

When you output the revised report, please follow the guidelines below to complete your task: \\
1. Carefully read and understand both the initial medical report and the retrieved relevant medical reports. \\
2. Identify any discrepancies, conflicts, or inconsistencies between the two reports. \\
3. Make necessary modifications to the initial medical report to resolve any identified issues, ensuring that the final report is clinically accurate and coherent. \\
4. If no changes are necessary, output the original initial medical report as is. \\

Here are two examples of the format you should output: \\
\{"action": "No", "report": "The region 0 is abdomen: ... The region 1 is bone: ... The region 2 is breast: ...", "reasons": ["reason1...", "reason2..."], "instructions": ["instruction1...", "instruction2..."]\}. \\
\{"action": "Yes", "report": "The region 0 is abdomen: ... The region 1 is bone: ... The region 2 is breast: ...", "reasons": ["reason1...", "reason2..."], "instructions": ["instruction1...", "instruction2..."]\}. 

Respond in JSON format without any additional content.
\end{prompt}

\begin{table}[thbp]
    \centering
    \setlength{\tabcolsep}{1pt}
    \resizebox{\linewidth}{!}{
    \begin{tabular}{lrr}
    \toprule
Characteristics & Train & Test\\
\midrule
Number of CT scans & 24,128 & 1,564 \\
Number of patients & 20,000 & 1,304 \\
Age (mean$\pm$std, years) & 48.74$\pm$17.28 & 48.39$\pm$16.87 \\
Sex (M/F) & 14,097/10,028 & 910/654 \\
\multicolumn{3}{l}{Regions} \\
~~Abdomen & 23,553 & 1,518\\
~~Bone & 23,479 & 1,509\\
~~Breast & 1,080 & 58\\
~~Heart & 23,289 & 1,433\\
~~Esophagus & 20,693 & 1,328\\
~~Lung & 23,741 & 1,514\\
~~Mediastinum & 23,684 & 1,523\\
~~Pleura & 18,156 & 1,172 \\ 
~~Thyroid & 1,093 & 51\\
~~Trachea/bronchi & 21,951 & 1,417\\
\multicolumn{3}{l}{Clinical Abnormalities} \\
~~Arterial wall calcification & 6,607 & 423 \\
~~Atelectasis & 6,005 & 359 \\
~~Bronchiectasis & 2,341 & 163 \\
~~Cardiomegaly & 2,533 & 159 \\
~~Consolidation & 4,066 & 280 \\
~~Coronary artery wall calcification & 5,747 & 348 \\
~~Emphysema & 4,558 & 304 \\
~~Hiatal hernia & 3,391 & 197 \\
~~Interlobular septal thickening & 1,887 & 119 \\
~~Lung nodule & 10,999 & 697 \\
~~Lung opacity & 8,944 & 607 \\
~~Lymphadenopathy & 5,839 & 343 \\
~~Medical material & 2,846 & 149 \\
~~Mosaic attenuation pattern & 1,788 & 124 \\
~~Peribronchial thickening & 2,566 & 178 \\
~~Pericardial effusion & 1,641 & 106 \\
~~Pleural effusion & 2,628 & 179 \\
~~Pulmonary fibrotic sequela & 6,175 & 399 \\
\bottomrule
    \end{tabular}
    }
    \caption{Statistics of the RadGenome-Chest CT dataset.}
    \label{tab:dataset}
\end{table}

\section{Statistics of Datasets}
\label{appendix:dataset}
In Table~\ref{tab:dataset}, we provide detailed statistics of the RadGenome-ChestCT dataset used in our experiments, including the number of samples, patient demographics, and the distribution of region-specific reports and clinical abnormalities. 

The dataset contains a total of 25,692 3D chest CT scans from 21,304 unique patients. Following the dataset's standard split, 24,128 scans are allocated for training and 1,564 for testing. Demographics are consistent across both sets, with a mean age of approximately 48.7 years ($\pm$17.2). The cohort includes 15,007 male and 10,682 female cases.
This dataset is de-identified and publicly available, and its use has been approved by the Institutional Review Board (IRB) of the institutions involved in its creation.

The dataset provides high-granularity reports across 10 anatomical regions, ensuring comprehensive spatial coverage. The most prevalent regions are the mediastinum, lung, and abdomen, each appearing in more than 25,000 scans. Additionally, the reports are annotated via RadBERT-RoBERTa-4m~\cite{yan2022radbert} with 18 distinct abnormalities of varying clinical prevalence. Lung nodules are the most frequent pathology ($n=11,696$), whereas cardiovascular findings such as arterial wall calcification ($n=7,030$) and pulmonary abnormalities such as lung opacity ($n=9,551$) provide a diverse range of diagnostic targets. 

\section{Case Study}
\label{appendix:case_study}
\begin{figure*}[htbp] 
    \centering    
    \includegraphics[width=0.8\textwidth]{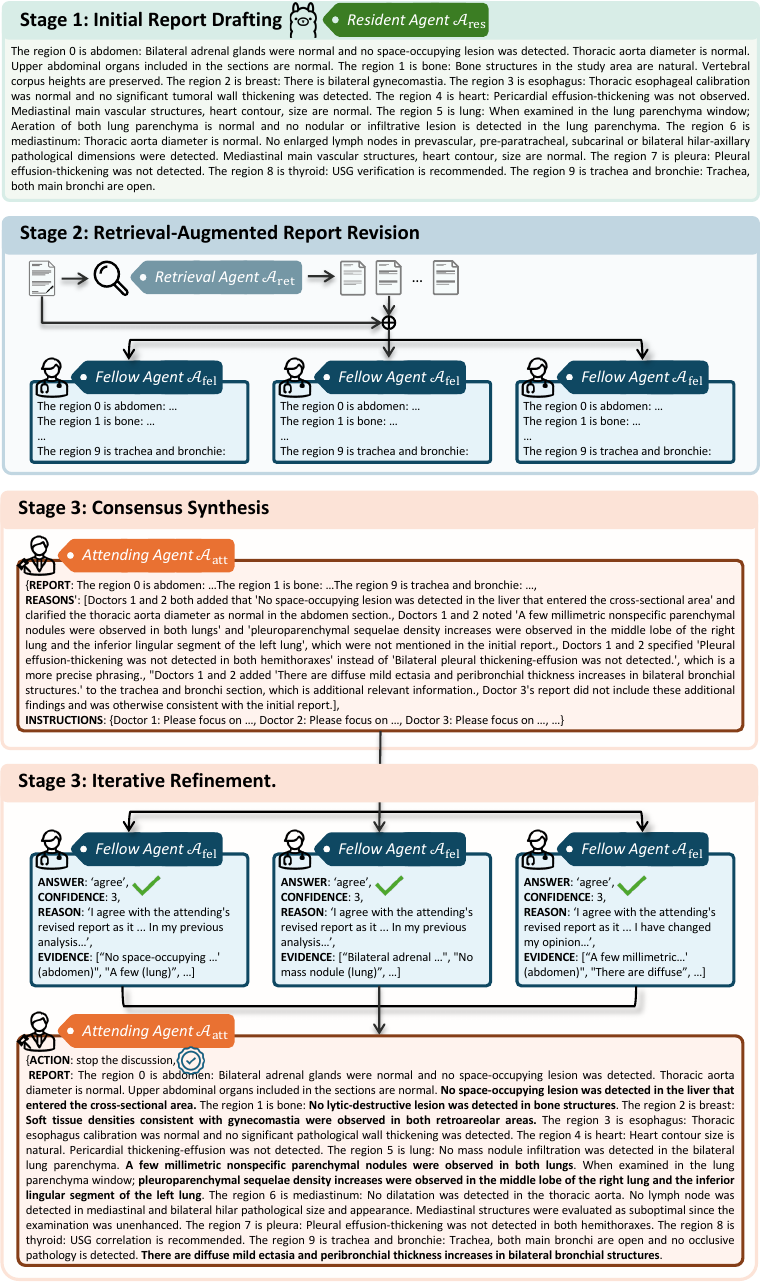}  
    \caption{Case study.}
    \label{fig:case_study} 
\end{figure*}
To demonstrate \ourmodel's capacity for generating interpretable, clinically grounded reports, we present a case study in Figure~\ref{fig:case_study}, which follows our hierarchical, multi-agent workflow and organizes findings by anatomical region.

\paragraph{Stage 1: Initial Report Drafting.} 
The Resident Agent $\mathcal{A}_{\text{res}}$ first generates a draft report directly from the CT images. It describes each predefined region. In this example, $\mathcal{A}_{\text{res}}$ reports largely normal findings across the abdomen (e.g., normal adrenal glands and upper abdominal organs), bones (preserved vertebral body heights), esophagus (no abnormal wall thickening), heart and mediastinum (normal contours and vessels, no effusion), lungs (no nodular/infiltrative lesion), pleura (no effusion/thickening), and airway (trachea and main bronchi patent), while also noting gynecomastia and recommending thyroid ultrasound correlation.

\paragraph{Stage 2: Retrieval-Augmented Report Revision.} 
Next, the Retrieval Agent $\mathcal{A}_{\text{ret}}$ query a reference database for visually and semantically similar studies using complementary strategies, including \textit{Image-to-Image retrieval}, \textit{Image-to-Text retrieval}, and \textit{Logits-based retrieval}.
Conditioned on this retrieved evidence, multiple Fellow Agents $\mathcal{A}_{\text{fel}}$ refine the draft. This step encourages diagnostic diversity and completeness.
In the shown case, fellows add or sharpen clinically relevant details (e.g., specifying no focal liver lesion within the imaged field, identifying a few millimetric nonspecific pulmonary nodules, describing pleuroparenchymal sequelae densities, refining pleural phrasing, and noting diffuse mild bronchial ectasia with peribronchial thickening).

\paragraph{Stage 3: Consensus-Driven Finalization.\\Round 1: Consensus Synthesis.}
The Attending Agent $\mathcal{A}_{\text{att}}$ subsequently consolidates these revisions into a single report and produces an explicit feedback report comprising: (i) key findings, (ii) diagnostic rationales that trace which fellows introduced each modification (e.g., additional lung and bronchial findings not present in the initial draft), and (iii) targeted suggestions to guide the next round of improvement.

\noindent\textbf{Round $t+1$: Iterative Refinement.}
Finally, fellows update their reports based on the attending's critique, specifically addressing localized findings such as bronchial wall thickening or hepatic textures. Each fellow articulates a formal stance, including agreement confidence, reasons,  and evidentiary support.
The attending adjudicates any remaining discrepancies, and once consensus is reached, terminates the discussion and releases the finalized report. In this example, the final report preserves the resident's normal baseline findings while incorporating consensus additions (e.g., nonspecific millimetric lung nodules, pleuroparenchymal sequelae, and bronchial ectasia/peribronchial thickening) and maintaining recommendations such as thyroid ultrasound correlation.

\section{Effectiveness of Agent Number}
\label{appendix:ablation_agent_num}
\begin{table}[t]
\small
\centering
\begin{tabular}{ccccc}
\toprule
Number & BLEU-1 & BLEU-4 & METEOR & CE-F1 \\
\midrule
1 & 0.473 & 0.253 & 0.451 & 0.323 \\
3 & 0.470 & 0.255 & 0.456 & 0.330 \\
5 & 0.476 & 0.257 & 0.455 & 0.335 \\
10 & 0.473 & 0.254 & 0.455 & 0.337 \\
20 & 0.475 & 0.255 & 0.454 & 0.327 \\
\bottomrule
\end{tabular}
\caption{Effectiveness of different \textit{Fellow Agent} $\mathcal{A}_{\text{fel}}$ numbers on the subset of RadGenome-ChestCT.}
\label{tab:abl_agent_num}
\end{table}
We conduct an ablation study to evaluate the sensitivity of \ourmodel to the number of participating Fellow Agents $\mathcal{A}_{\text{fel}}$. As detailed in Table~\ref{tab:abl_agent_num}, we vary the agent count from 1 to 20. Due to the budget constraints for LLM usage, the evaluation is conducted on a subset of 100 samples from the RadGenome-ChestCT test set.

The experimental results indicate that increasing the number of Fellow Agents initially enhances report quality. An ensemble of 5 agents achieved the highest linguistic quality, reaching a BLEU-1 of 0.476 and a METEOR score of 0.455. 10-agent ensembles slightly improved the F1 score to 0.337. However, increasing the ensemble size to 20 agents led to a marginal decline in performance, suggesting that excessive agent density may introduce redundant information or discursive noise. These findings indicate that a moderate number of agents effectively balance diverse diagnostic perspectives with collaborative stability. To optimize the trade-off between report fidelity and inference cost, we utilized 3 Fellow Agents as the default configuration for experiments on the whole dataset.

\begin{figure*}[thbp] 
    \centering    
    \includegraphics[width=\textwidth]{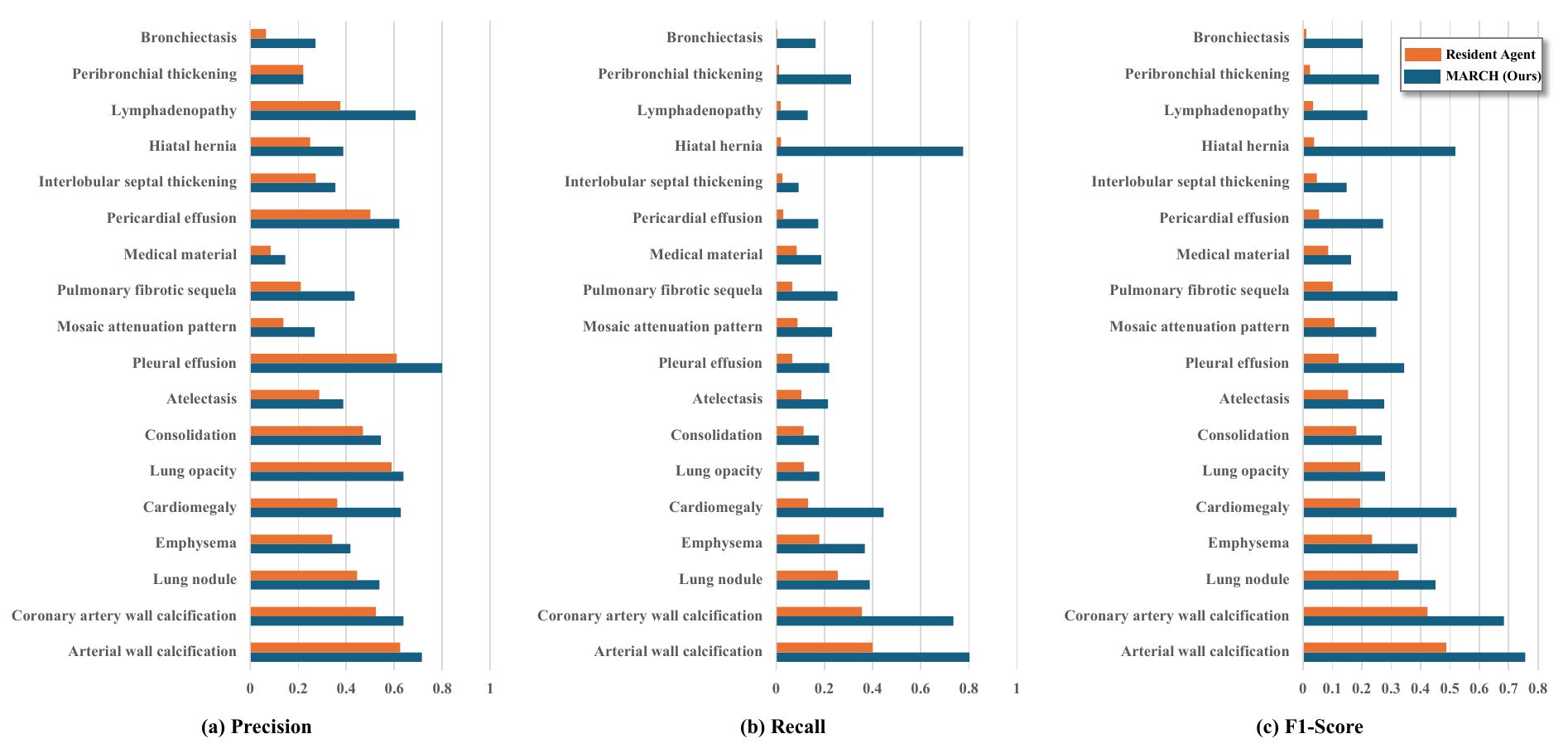}  
    \caption{Clinical efficacy in terms of (a) Precision, (b) Recall, and (c) F1-score.}
    \label{fig:class} 
\end{figure*}
\begin{figure*}[thbp] 
    \centering    
    \includegraphics[width=\textwidth]{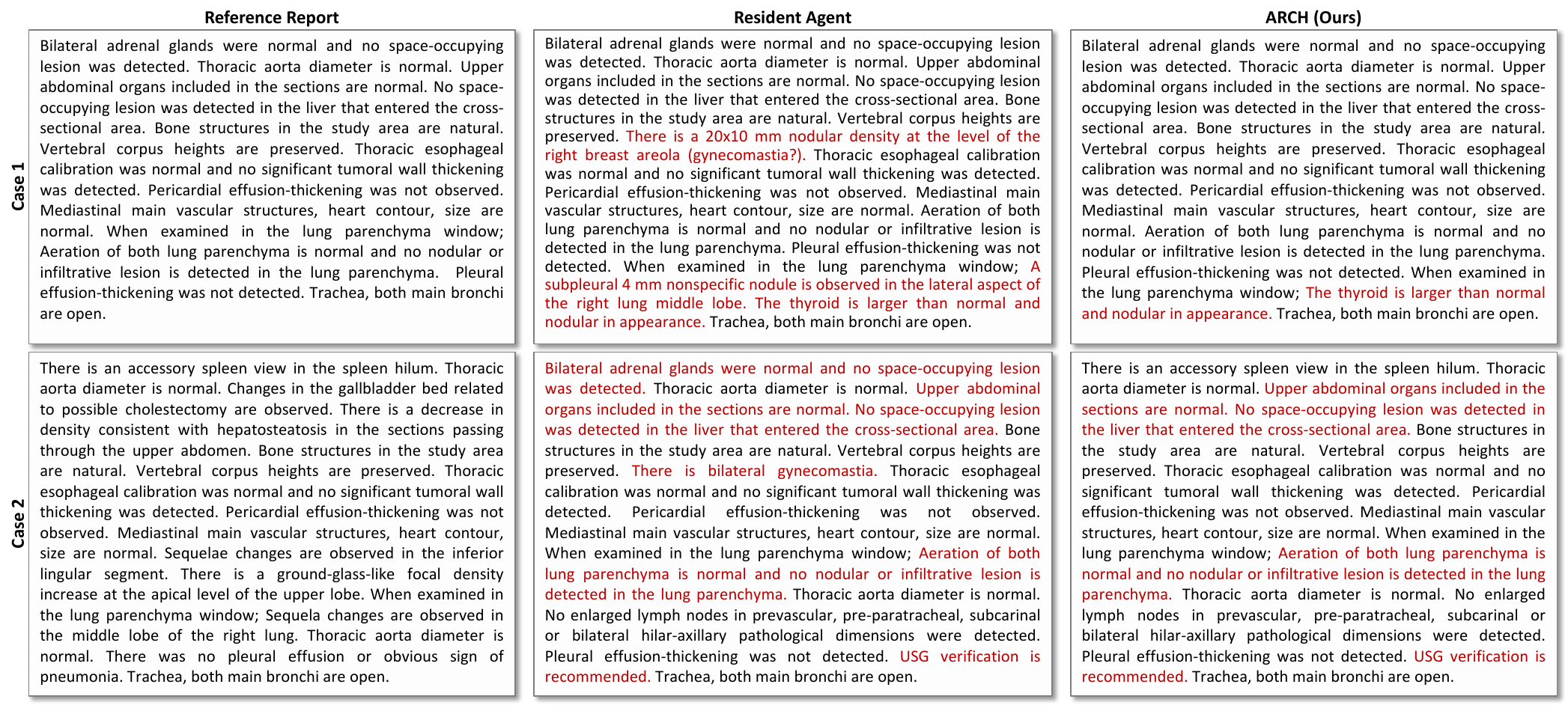}  
    \caption{Examples of the generated reports.}
    \label{fig:example} 
\end{figure*}

\section{Clinical Efficacy across Clinical Abnormalities}
\label{appendix:clinical_efficacy}
In Figure~\ref{fig:class}, we present a detailed analysis of \ourmodel's clinical efficacy across 18 clinical abnormalities in the RadGenome-ChestCT dataset, including \textit{arterial wall calcification, coronary artery wall calcification, lung nodule, emphysema, cardiomegaly, lung opacity, consolidation, atelectasis, pleural effusion, mosaic attenuation pattern, pulmonary fibrotic sequela, medical material, pericardial effusion, interlobular septal thickening, hiatal hernia, lymphadenopathy, peribronchial thickening, and bronchiectasis}.
We compare \ourmodel against the Resident Agent $\mathcal{A}_{\text{res}}$ baseline that generates reports without multi-agent collaboration.
Results demonstrate that \ourmodel consistently outperforms the baseline across all abnormalities in terms of Precision, Recall, and F1-Score. Specifically, \ourmodel achieves high recall for abnormalities such as Hiatal hernia, Coronary artery wall calcification, and Arterial wall calcification, with scores exceeding 0.8. In terms of the overall F1-Score, \ourmodel shows significant gains in identifying complex findings such as Arterial wall calcification, Coronary artery wall calcification, and Cardiomegaly, while maintaining a robust balance between precision and sensitivity.

\section{Examples of the Generated Report}
\label{appendix:examples}
In Figure~\ref{fig:example}, we provide examples of radiology reports generated by \ourmodel compared to those produced by the Resident Agent $\mathcal{A}_{\text{res}}$ baseline. 
$\mathcal{A}_{\text{res}}$'s initial drafts often include extraneous or uncertain observations, such as suspected gynecomastia or small nonspecific nodules.
In contrast, \ourmodel delivers refined reports that more closely align with the reference reports by effectively filtering these potential hallucinations and emphasizing clinically relevant findings through hierarchical collaboration.
For instance, \ourmodel accurately identifies the thyroid's enlarged and nodular appearance while maintaining a concise summary of normal findings in the abdomen and bone structures. In Case 2, while the Resident Agent focuses on standard anatomical observations, \ourmodel demonstrates superior clinical fidelity by correctly identifying specific anatomical variants, such as an accessory spleen, and incorporating critical recommendations for USG verification.
These examples underscore \ourmodel's capability to leverage multi-agent collaboration to resolve diagnostic ambiguities and generate more reliable, clinically validated radiology reports.
\end{document}